\documentclass[conference]{IEEEtran}
\IEEEoverridecommandlockouts
\usepackage{amsmath,amssymb,amsfonts}
\usepackage{algorithmic}
\usepackage{graphicx}
\usepackage{textcomp}
\usepackage{amsfonts}
\usepackage[ruled,vlined]{algorithm2e}
\usepackage{booktabs}
\usepackage{cite}

\usepackage[usernames,dvipsnames]{xcolor}

\def\BibTeX{{\rm B\kern-.05em{\sc i\kern-.025em b}\kern-.08em
    T\kern-.1667em\lower.7ex\hbox{E}\kern-.125emX}}
\begin{document}

\title{Modeling, Characterization, and Control of Bacteria-inspired Bi-flagellated Mechanism with Tumbling\\
\thanks{
We acknowledge financial support from the National Science Foundation under Grant numbers CAREER-2047663 and CMMI-2101751.}
}

\author{Zhuonan Hao$^{1}$,
Sangmin Lim$^{1}$, Mohammad Khalid Jawed$^{1}$
\thanks{$^{1}$Department of Mechanical \& Aerospace Engineering, University of California, Los Angeles, 420 Westwood Plaza, Los Angeles, CA 90095}
}

\maketitle

\begin{abstract}
Multi-flagellated bacteria utilize the hydrodynamic interaction between their filamentary tails, known as flagella, to swim and change their swimming direction in low Reynolds number flow. This interaction, referred to as bundling and tumbling, is often overlooked in simplified hydrodynamic models such as Resistive Force Theories (RFT). However, for the development of efficient and steerable robots inspired by bacteria, it becomes crucial to exploit this interaction. In this paper, we present the construction of a macroscopic bio-inspired robot featuring two rigid flagella arranged as right-handed helices, along with a cylindrical head. By rotating the flagella in opposite directions, the robot's body can reorient itself through repeatable and controllable tumbling. To accurately model this bi-flagellated mechanism in low Reynolds flow, we employ a coupling of rigid body dynamics and the method of Regularized Stokeslet Segments (RSS). Unlike RFT, RSS takes into account the hydrodynamic interaction between distant filamentary structures. Furthermore, we delve into the exploration of the parameter space to optimize the propulsion and torque of the system. To achieve the desired reorientation of the robot, we propose a tumble control scheme that involves modulating the rotation direction and speed of the two flagella. By implementing this scheme, the robot can effectively reorient itself to attain the desired attitude. Notably, the overall scheme boasts a simplified design and control as it only requires two control inputs. With our macroscopic framework serving as a foundation, we envision the eventual miniaturization of this technology to construct mobile and controllable micro-scale bacterial robots.

\end{abstract}



\section{Introduction}
\label{intro}

The study of flagellated bacteria and microorganisms has provided valuable insights into the development of flagellated robots.~\cite{brennen1977fluid,lauga2009hydrodynamics, dolev2021board}. 
These robots mimic the locomotion of flagellated organisms, which rely on the intricate interaction between their helical structures, known as flagella, and the surrounding viscous fluid. By understanding and replicating these propulsion mechanisms, flagellated robots can achieve functional movements such as running, turning, and stopping. Additionally, natural observations have highlighted that different types of bacteria, including uni-flagellated and multi-flagellated species, rely on distinct propulsion mechanisms to achieve specific forms of locomotion \cite{dvoriashyna2021hydrodynamics, taylor1951analysis}.

Uni-flagellated bacteria possess a single flagellum filament protruding from the side of the body, enabling them to swim through the rotary motion of the flagellum relative to the cell body \cite{lim2022fabrication, taylor1951analysis}. Notably, previous investigations have revealed that when the rotational frequency of the flagella exceeds a threshold, buckling instability occurs, resulting in highly nonlinear swimming trajectories \cite{vogel2012motor, jawed2015propulsion}. The body orientation of these robots can be controlled simply by adjusting the spin speed of the flagellum, a mechanism that has been widely employed in uni-flagellated robot designs to achieve motility \cite{du2022modeling, son2013bacteria}. However, research on robotic propulsion inspired by multi-flagellated bacteria is relatively limited. 

Multi-flagellated bacteria exhibit locomotion through the interplay of their flagella, involving phenomena such as bundle formation, tumbling, and polymorphic transformations, all of which arise from different flagella actuation \cite{lim2022fabrication, berg2003rotary, berg1972chemotaxis}. Bundle formation occurs when two or more flagella spin in the same direction, generating efficient longitudinal propulsion. The propulsive force is approximately linearly related to the spin speed, and this observation has been effectively utilized in bi-flagellated robots to enable single-direction mobility \cite{lim2021bacteria, ye2013rotating}. The presence of multiple flagella in these robots offers benefits, suggesting alternative approaches for speed enhancement beyond flagellum geometry optimization. However, a major limitation arises in the area of turning or reorienting the body, preventing these robots from swimming freely in space. Recent research has explored changing the spin direction of one or more flagella, gradually reducing the propulsion thrust and generating a turnover torque \cite{dvoriashyna2021hydrodynamics}. This results in rapid tumble events and seemingly erratic body reorientation. Our study models the tumbling event as a predictable phenomenon and aims to incorporate the tumbling mechanism into a bi-flagellated robot to enhance steerability.

\begin{figure}[b!]
    \centering
    \includegraphics[width=\linewidth]{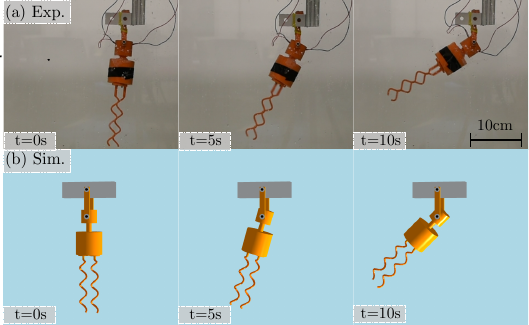}
    \caption{Snapshots from (a) experiments and (b) simulations. Side view of the bi-flagellated rotating around an axis at $t \in \{0,5,10\} \text{s}$. Two identical flagella rotate in opposite directions, i.e., clockwise (CW) and counterclockwise (CCW), at an angular velocity of $\omega = 280~\text{rpm}$.}
    \label{fig:exp_sim_overview}
\end{figure}

To imitate the fluid-structure interplay between flagellum and low Reynolds number flow, computational fluid dynamics model including Resistive Force Theory (RFT), Slender Body Theory (SBT) \cite{lighthill1976flagellar}, and Regularized Stokeslets Segments (RSS) \cite{cortez2018regularized} are used to predict the motion of uni- and multi-flagellated robots. RFT introduces the drag coefficient along the tangential and perpendicular directions of the flagellum. The method is computationally inexpensive but neglects the hydrodynamic interactions between flows induced by different parts of the flagellum. An accurate quantitative analysis requires a non-local hydrodynamics force model that accounts for the interaction between the flow induced by distant parts of the filament. Both SBT and RSS rely on the linearity of the Stokes equations for low Reynolds number flow, which can accurately describe the evolution of flagellum dynamics with long-ranged interactions in a viscous fluid.

To understand the physical phenomenon of flagellated locomotion, we couple the rigid body dynamics with the hydrodynamics model to simulate the robot's trajectory. Position and orientation are observable states in the system, which allow us to study periodical locomotion. As shown in Figure \ref{fig:exp_sim_overview}, the experiment and simulation show good agreement. The proposed simulation framework successfully reveals the interaction between two flagella from long-ranged hydrodynamics.

Our contributions are as follows. We model and create a macroscopic bi-flagellated robot to study how different actuation modes can switch robot locomotion patterns. A simple bi-flagellated robot that exploits variation in viscosity and structure in its tails can effectively reorient the body. A framework comprising experiments and simulations is developed to study the robot's locomotion. The simulation tool can be used to generate data to explore the parameter space of the tumbling phenomenon. Meanwhile, the robot's dynamics are fully described and can be used to formulate the control scheme. The physics behind the tumbling locomotion is elaborated in detail. The simplicity of the robot and the small number of moving parts can eventually lead to the miniaturization of this robot.

The paper is organized as follows. In section \ref{sec:experiment}, we demonstrate the structural design and experimental setup of the bi-flagellated robot. Section \ref{sec:method} presents a computational framework to describe the robot dynamics in a viscous fluid. Section \ref{sec:result} explores the optimal robot geometry for best dynamics performance, and we validate our simulation results against the experiments. Section \ref{sec:conclusion} concludes our works and proposes the potential directions for future study.

\section{Experimental design}
\label{sec:experiment}
\subsection{Robotic structure}

\begin{figure}[ht!]
    \centering
    \includegraphics[width=\linewidth]{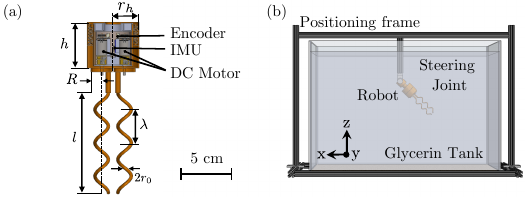}
    \caption{Robot schematic and experimental setup. (a) The bio-inspired robot is comprised of two components: (i) a cylindrical head with radius $r_h$ and height $h$, and (ii) two helical flagella tails with radius $r$, length $l$, pitch $\lambda$, and cross-section radius $r_0$. (b) The bi-flagellated robot immerses into glycerin and rotates its body around a steering joint (1-DOF), i.e., y-axis. The rotations around other axes and translations are limited for. }
    \label{fig:design_setup}
\end{figure}

The robot depicted in Figure \ref{fig:design_setup} (a) consists of a cylindrical head and two right-handed helical flagella with plates attached to the motor shaft. The head has a radius of $r_h$ measuring 2.5 cm and a height of $h$ measuring 4.3 cm. Inside the head, two tiny brushed DC geared motors are located, each rated at 6 V voltage and capable of a stall current of 1.5 A. The motors are equipped with magnetic encoders and an IMU module. The motor shaft protrudes from the head, and its rotation direction and speed are controlled via PWM feed using a microcontroller. The flagella are manufactured using rapid prototyping techniques with Polylactic acid (PLA), a type of 3D printing material. The PLA flagella have a fixed cross-sectional radius of $r_0$ measuring 1.58 mm and a helix radius $R$ measuring 6.36 mm. To generate sufficient experimental data for investigating the tumbling mechanism, the helix length $l$ is varied between 63.6 and 127.2 mm, while the helix pitch $\lambda$ is varied between 15.9 and 63.6 mm. The PLA material used for the flagella is considered to be non-deformable, with a Young's modulus $E$ of 4.107$\times10^9$ Pa. These design and parameter variations allow for the exploration of different configurations of the flagellated robot and provide a range of experimental data to study the underlying mechanisms of tumbling.

\subsection{Experimental setup}
Our experiments are designed to serve two main purposes: (i) to explore the optimal structure of the flagellum and robot that generates maximum steering efficiency, and (ii) to achieve controllable direction changes. The details of our experimental setup are as follows.

Figure \ref{fig:design_setup} (b) illustrates the experimental apparatus used to validate the numerical simulations developed in the subsequent section. The platform comprises four components: (i) a glycerin tank with dimensions of 122 cm (length) $\times$ 45 cm (width) $\times$ 51.5 cm (height), (ii) the bi-flagellated robot, (iii) a steering joint, and (iv) a positioning frame. Glycerin, with a density $\rho$ of 1.26 g/ml and viscosity $\mu$ of 1 Pa$\cdot$s at 25$^\circ$ Celsius, is chosen as the surrounding viscous environment. To facilitate quantitative comparison, we restrict the robot's degree of freedom (DOF) to a single-axis rotation using the steering joint. This setup enables us to characterize the tumbling behaviors associated with different flagellum designs. The DC geared motors located inside the robot's head are connected to an external microcontroller, allowing us to adjust the rotational speed and direction. The test platform permits us to vary the rotation speed $\omega$ within the range of 0 to 30 rad/s, ensuring that the low Reynolds number condition is satisfied, i.e., $Re = \rho \omega R r_0/\mu \leq 0.37$. This condition guarantees that the fluid flow is predominately governed by viscosity rather than inertia. By operating within the low Reynolds number regime, we can accurately investigate the hydrodynamic interactions between the flagella and the surrounding fluid.


\section{Numerical method of bi-flagellated locomotion}
\label{sec:method}

\begin{table}[ht]
  \centering
  \caption{Parameters with symbolic representations}
  \label{tab:para}
  \begin{tabular}{ cccc }
    \toprule
    Symbol & Value & Unit & Description \\  
    \midrule
    $E$ & 4.107$\times10^9$ & Pa & Young's modulus \\ 
    $\rho$ & 1260 & kg/$\mathrm{m}^{3}$  & Density \\    
    $\mu$ & 1 & Pa$\cdot$s & Viscosity \\ 
    \midrule
    $r_0$ & 1.58 & mm & Cross-sectional radius \\
    $R$ & 6.36 & mm & Helix radius \\  
    $\lambda$ & 31.8 & mm & Helix pitch \\  
    $l$ & 95.4 & mm & Helix axial length \\
    $d$ & 22 & mm & Flagellum spacing distance \\ 
    \midrule
    $r_h$ & 25 & mm & Head radius \\ 
    $h$ & 43 & mm & Head height \\ 
    $r_m$ & 2 & mm & Mass center shift \\
    \midrule
    $m_h$  & 0.1 & kg & Head mass \\  
    $g$  & 9.8 & m/$\mathrm{s}^{2}$ & Gravitational acceleration \\  
    $\omega$  & 0.1 & rad/s & Flagellum rotation speed \\
    \midrule
    $\Delta l$  & 5 & mm & Discretization length \\ 
    $\Delta t$  & 0.01 & s & Time step \\ 
    $C_t$ & 4 & 1 & Translational drag coefficient \\
    $C_r$ & 2.3 & 1 & Rotational drag coefficient \\
    \bottomrule
  \end{tabular}
\end{table}
The bi-flagellated robot consists of two main components: the helical flagellum and the cylindrical head. In order to simulate the locomotion of this robot in a viscous fluid medium, we develop a numerical model that combines three key components: (i) a kinematic representation for the bi-flagellated mechanism, (ii) Regularized Stokeslet Segments (RSS) to model the long-ranged hydrodynamics forces, and (iii) a forced-head hydrodynamics model to capture the interaction between the flagellum and the head. This section is structured as follows. In Section \ref{sec:kinematic}, we provide a description of the kinematic representation of the helical flagellum and cylindrical head. This representation allows us to characterize the motion of the robot in terms of its shape and orientation. Then, in Section \ref{sec:RSS}, we explain how we integrate the kinematic representation with the RSS method to compute the hydrodynamic forces exerted on the flagellum. The RSS method considers the interactions between different segments of the flagellum and the surrounding fluid. Next, in Section \ref{sec:forced-head}, we detail the equations of motion (EOM) that govern the dynamics of the bi-flagellated robot. These EOM are derived from the theory of rigid body dynamics, taking into account the forces and torques acting on the flagellum and the head. By solving these equations, we can simulate the motion and behavior of the robot in response to the hydrodynamic forces. Finally, in Section \ref{sec:definition}, we discuss the geometric and physical conditions of the problem, including the dimensions and material properties of the flagellum and head. These conditions play a crucial role in determining the behavior and performance of the bi-flagellated robot in the simulated fluid environment.

\subsection{Kinematic representation}
\label{sec:kinematic}
\subsubsection{Helical flagellum} We model the flagella filament as a perfect helix with radius $R$, pitch $\lambda$, and axial length $l$ (see Figure \ref{fig:design_setup} (a)). A right-handed helix in the Cartesian coordinate system is parameterized as a function of $s$, i.e.,
\begin{align}
\label{equ:helix_generation}
    \mathbf{r}(s) = \left[R\cos \frac{2\pi s}{L} , R \sin \frac{2\pi s}{L},  \frac{\lambda s}{L} \right], 0 \leq s \leq l,
\end{align}
where $L = \sqrt{(2\pi R)^2 + \lambda^2}$ is the contour length of one helical turn.
In the case of a left-handed helix, the second term has a negative sign. We employ the discretization method to model the kinematic of helical filament. In the schematic of Figure \ref{fig:RSS} (a), each discrete helical curve consists of $N+1$ nodes, i.e., $\mathbf{n} = [\mathbf{n}_0, \mathbf{n}_1, \cdots, \mathbf{n}_{N}]$. We take the first two nodes $\mathbf{n}_0$ and $\mathbf{n}_1$ as the connection to the head. Starting from $i = 2$, the coordinate of the remaining nodes is calculated by taking $s = (i-2)l/(N-2)$ in Equation \ref{equ:helix_generation}. 

The $N+1$ nodes correspond to $N$ edge vector $\mathbf{e}^1$, $\cdots$,$\mathbf{e}^{N-1}$, such that $\mathbf{e}^i = \mathbf{x}_{i+1}-\mathbf{x}_i$, where $i = 1, \cdots, N-1$. Hereby, we denote the node-associated quantities by superscripts and edge-associated quantities by subscripts. Nodal positions constitute the $3N$ sized DOF vector, i.e., $\mathbf{X} = [\mathbf{x}_0, \cdots, \mathbf{x}_N]^T$, where the superscripts $T$ denotes transposition. 

Because the rigid flagellum can only rotate around a single fixed axis, i.e., z-axis, the angular velocity vector is specified as $\boldsymbol{\omega} = [0, 0, \omega_z]$. By defining the rotation axis $\mathbf{x}_{\text{rotate}} = [0,0,1] $, we can obtain the linear velocity of each node by $\dot{\mathbf{x}}_i = \boldsymbol{\omega} \times (\mathbf{x}_i-\mathbf{x}_{\text{rotate}})$, where $\times$ denotes the cross product of two vectors. With nodal velocities, we can update the nodal positions at each time step. Rearranging the derivative of DOF vector as $\mathbf{U} = \dot{\mathbf{X}} = [\dot{\mathbf{x}}_0, \cdots, \dot{\mathbf{x}}_N]^T$, the variable is used to formulate the drag force in RSS (see details in Section \ref{sec:RSS}).

\subsubsection{Cylindrical head} Without losing the generality, we use a single head node $\mathbf{n}_{\text{head}}$ in Figure \ref{fig:RSS} (a.1) to represent the spatial configuration of the bi-flagellated system. Concerning a prescribed fixed coordinate system, denoted as inertial frame $\mathbf{x}^I: x^I-y^I-z^I$, we can describe the translation and rotation on the rotated coordinate system, designated as body frame $\mathbf{x}^B: x^B-y^B-z^B$ attached to the head (see Figure \ref{fig:RSS} (b)). We take the Euler angles to represent the orientation of the head, which are typically denoted as yaw $\alpha$, pitch $\beta$, and roll $\gamma$ in $Z-Y-X$ convection. In the steering joint setup, we define pitch angle $\beta$ as the angle between the axis $z_I$ and axis $z_B$ to describe the orientation of the bi-flagellated system. Further, to model the free swimming motion after removing the steering joint, we introduce quaternion for orientation $\mathbf{q} = (q_0,q_1,q_2,q_3)$ (convertable to Euler angle) and axial angular velocities along body frame $\boldsymbol{\omega}^B =  (\omega_x^B,\omega_y^B,\omega_z^B)$, and define DOF vector $\mathbf{Q} = [\mathbf{x}^{I},\dot{\mathbf{x}}^{I},\mathbf{q},\boldsymbol{\omega}^{B}]$ to represent spatial information.

\subsection{Regularized Stokeslets Segments}
\label{sec:RSS}
\begin{figure}[ht!]
    \centering
    \includegraphics[width=\linewidth]{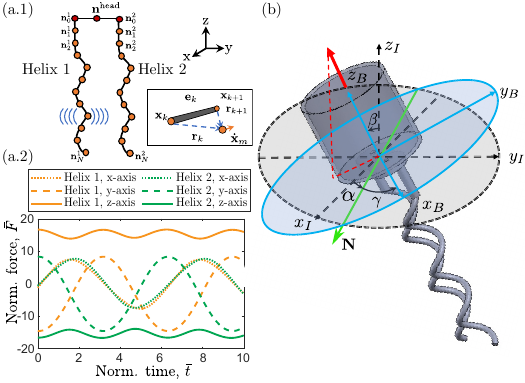}
    \caption{Kinematic representation of bi-flagellated system. (a) Long-range hydrodynamics. (a.1) Discrete schematic of the bi-flagellated robot. Each helical flagellum is discretized into $N+1$ nodes. Superscript denotes the helix index, and subscript denotes the node index. $\mathbf{n}_0^1$ and $\mathbf{n}_0^2$ connect the flagellum with the head node, as the application points of the forces generated by each helix, interacting with the head node in rigid body dynamics. Inset: Notations associated with the flow $\dot{\mathbf{x}}_m$ at point $\mathbf{x}_m$ generated by a line segment $\mathbf{e}_k = \mathbf{x}_{k+1}-\mathbf{x}_{k}$. (a.2) Time series of normalized hydrodynamics forces $\Bar{F}$ along x, y, and z direction with flagella spacing distance $d = 0.022 \text{ m}$ and rotation speed $\omega_1=1 \text{ rad/s}$ and $\omega_2=-1 \text{ rad/s}$. The sinusoidal wave pattern arises due to the long-ranged hydrodynamics interactions between flows induced by different segments of the flagella. (b) The description of the robot coordinate in the body frame and inertia frame with Euler's angle representation. The green arrow line $\mathbf{N}$ indicate the line of nodes.}
    \label{fig:RSS}
\end{figure}

We use Regularized Stokeslets Segments (RSS) methods to model the viscous drag force experienced by a helical flagellum in motion within a viscous fluid. The relation between the velocity vector $\mathbf{U}$ of (size $3N$) at nodes set $\mathbf{n}$ and the hydrodynamics force vector $\mathbf{F}$ of (size $3N$) applied on them is linearly configured by a geometry-associated matrix $\mathbf{A}$ (vector of size $3N\times3N$), i.e., 
\begin{align}
\label{equ:linear_relation}
    \mathbf{U} = \mathbf{A}\mathbf{F},
\end{align}
We describe the formulation of matrix $\mathbf{A}$ on the discretized helical flagellum as follows.

The primary Green's function of Stokes flow is the Stokeslets, which describes the flow associated with a singular point force. Referring to Figure \ref{fig:RSS} (a), RSS provides a relationship between the velocity $\dot{\mathbf{x}}_m$ at a point $\dot{\mathbf{x}}_m$ and the forces applied by each node on the fluid such that
\begin{equation}
8 \pi \mu \dot{\mathbf{x}}_m=\sum_{k=0}^{N-2}\left(\mathbf{A}_1^k \mathbf{f}_{\mathrm{k}}^h+\mathbf{A}_2^k \mathbf{f}_{\mathrm{k}+1}^h\right),
\end{equation}
where $\mathbf{f}_{\mathrm{k}}$ is the force vector of size 3 that represents the force applied by the $k$-th node onto the fluid. This is equal and opposite to the hydrodynamics force onto the $k$-th node. The matrix $\mathbf{A_1^k}$ and $\mathbf{A_2^k}$ are
\begin{equation}
\begin{array}{r}
\mathbf{A}_2^k=\left|\mathbf{s}_k\right|\left(\left(T_{1,-1}^{k, k+1}+c^2 T_{1,-3}^{k, k+1}\right) \mathbf{I}+T_{1,-3}^{k, k+1}\left(\mathbf{r}_k \mathbf{r}_k^T\right)+\right. \\
\left.T_{2,-3}^{k, k+1}\left(\mathbf{r}_k \mathbf{s}_k^T+\mathbf{s}_k \mathbf{r}_k^T\right)+T_{3,-3}^{k,-3}\left(\mathbf{s}_k \mathbf{s}_k^T\right)\right), \\
\mathbf{A}_1^k=\left|\mathbf{s}_k\right|\left(\left(T_{0,-1}^{\mathbf{k}, \mathbf{k}+1}+c^2 T_{0,-3}^{\mathbf{k}, \mathbf{k}+1}\right) \mathbf{I}+T_{0,-3}^{k, k+1}\left(\mathbf{r}_k \mathbf{r}_k^T\right)+\right. \\
\left.T_{1,-3}^{k, k+1}\left(\mathbf{r}_k \mathbf{s}_k^T+\mathbf{s}_k \mathbf{r}_k^T\right)+T_{2,-3}^{k, k+1}\left(\mathbf{s}_k \mathbf{s}_k^T\right)\right)-\mathbf{A}_2^k,
\end{array} \notag
\end{equation}
where $\mathbf{x}_m$ is the point of measurement, $c$ is the regularization parameter (from analysis in \cite{cortez2018regularized}, $c = 1.031\cdot r_0$), $\mathbf{I}$ is the 3-b-3 identity matrix, $\mathbf{r}_k = \mathbf{x}_m - \mathbf{x}_k$, $\mathbf{r}_{k+1} = \mathbf{x}_m - \mathbf{x}_{k+1}$, and $\mathbf{e}_k = \mathbf{x}_{k+1} - \mathbf{x}_k$ are the position vectors between edge and point of measurement, and the scalar quantities denoted by $T$, e.g., $T_{0,-1}^{k,k+1}$ are expressed as follow
\begin{equation}
\begin{aligned}
&T_{0,-1}^{k, k+1}=\left.\frac{1}{\left|\mathbf{s}_k\right|} \log \left[\left|\mathbf{s}_k\right| R+\left(\mathbf{x}_\alpha \cdot \mathbf{s}_k\right)\right]\right|_0 ^1,\\
&T_{0,-3}^{k, k+1}=-\left.\frac{1}{R\left[\left|\mathbf{s}_k\right| R+\left(\mathbf{x}_\alpha \cdot \mathbf{s}_k\right)\right]}\right|_0 ^1,\\
&T_{1,-1}^{k, k+1}=\left.\frac{R}{\left(\left|\mathbf{s}_k\right|\right)^2}\right|_0 ^1-\frac{\left(\mathbf{x}_0 \cdot \mathbf{s}_k\right)}{\left(\left|\mathbf{s}_k\right|\right)^2} T_{0,-1}^{k, k+1} \text {, }\\
&T_{1,-3}^{k, k+1}=-\left.\frac{1}{R\left(\left|\mathbf{s}_k\right|\right)^2}\right|_0 ^1-\frac{\left(\mathbf{x}_0 \cdot \mathbf{s}_k\right)}{\left(\left|\mathbf{s}_k\right|\right)^2} T_{0,-3}^{k, k+1} \text {, } \\
&T_{2,-3}^{k, k+1}=-\left.\frac{\alpha}{R\left(\left|\mathbf{s}_k\right|\right)^2}\right|_0 ^1+\frac{1}{\left(\left|\mathbf{s}_k\right|\right)^2} T_{0,-1}^{k, k+1}-\frac{\left(\mathbf{x}_0 \cdot \mathbf{s}_k\right)}{\left(\left|\mathbf{s}_k\right|\right)^2} T_{1,-3}^{k, k+1},\\
&T_{3,-3}^{k, k+1}=-\left.\frac{\alpha^2}{R\left(\left|\mathbf{s}_k\right|\right)^2}\right|_0 ^1+\frac{2}{\left(\left|\mathbf{s}_k\right|\right)^2} T_{1,-1}^{k, k+1}-\frac{\left(\mathbf{x}_0 \cdot \mathbf{s}_k\right)}{\left(\left|\mathbf{s}_k\right|\right)^2} T_{2,-3}^{k, k+1},\notag
\end{aligned}
\end{equation}
where $\mathbf{x}_{\alpha} = \mathbf{x}_{k} - \alpha \mathbf{s}_k$, and $R = \sqrt{|\mathbf{x}_{\alpha}|^2 + c^2}$.

The geometry matrix $\mathbf{A}$ is formulated by a rearrangement of block matrices $\mathbf{A}_1^k$ and $\mathbf{A}_2^k$ on the corresponding node index. At each time step in the simulation, knowing the position $\mathbf{X}$ and velocity $\mathbf{U}$ of the node set $\mathbf{n}$, we can construct the geometry matrix $\mathbf{A}$. Then we employ the force-velocity relationship in Equation \ref{equ:linear_relation} to evaluate the hydrodynamics forces by $\mathbf{\hat{F}} = \mathbf{A} \setminus \mathbf{U}$.

\subsection{Forced-head hydrodynamics model}
\label{sec:forced-head}

To study the locomotion of a bi-flagellated system under external forces, we build the forced-head hydrodynamics model by using the aforementioned kinematic representation. A single head node $\mathbf{n}^{\text{head}}$ and two interconnected nodes from helical flagellum $\mathbf{n}_{0}^1$ and $\mathbf{n}_{0}^2$. in Figure \ref{fig:RSS} describe the configuration of the dynamical system. Head node $\mathbf{n}_{\text{head}}$ accounts for the hydrodynamics drag force and torque induced by the translation and rotation of the head. Two connected nodes $\mathbf{n}_{0}^1$ and $\mathbf{n}_{0}^2$ with equal distance $d/2$ to $\mathbf{n}^{\text{head}}$ account for the resultant of hydrodynamics forces and torque generated by two helical flagella. In this section, we analyze the forces and torques applied to the system and formulate the equation of motion of bi-flagellated locomotion. 

\textbf{Hydrodynamics force on head.} We use the Stokes law to compute the hydrodynamics force on the robot head. As the head translates with velocity $\dot{\mathbf{x}}^{\text{head}}$, the viscous fluid exerts a drag force to resist the translation. Likewise, when the head rotates with angular velocity $\boldsymbol{\omega}^\text{head}$, the viscous fluid applies torque to resist that rotation. Stokes' law model the hydrodynamics drag by considering the object as a small sphere. For the cylindrical head, we can introduce two prefactors $C_t, C_r$ to account for the non-sphericity of the object. Therefore, we model the translation drag force as
\begin{equation}
    \mathbf{f}_{t}^{\text {head }}=-C_t \cdot 6 \pi \mu r_h \dot{\mathbf{x}}^\text{head},
\end{equation}
where $r_h$ is the radius of the head, and the rotation drag torque as
\begin{equation}
    \mathbf{T}_r^{\text{head}} = -C_r \cdot 8 \pi d_r^3 \boldsymbol{\omega}^\text{head},
\end{equation}
where $d_r$ is reference dimension to account for the non-spherical shape. The value of $C_t$ and $C_r$ are determined by drop and rotation test. In our model, $\mathbf{f}_{t}^{\text{head }}$ and $\mathbf{T}_r^{\text{head}}$ are applied on the head node $\mathbf{n}^{\text{head}}$.

\textbf{Righting moment due to mass distribution.} In our bi-flagellated system, the head is conditioned to be neutral buoyancy. Therefore, the gravitational force and buoyancy forces are balanced, i.e., $mg = \rho V g$, where $m$ is the mass of the head, $V$ is the volume of the head, $\rho$ is the density of fluid medium, and $g$ is gravitational acceleration. However, the mass is not uniformly distributed along the robot head. When the center of mass (COM) and center of geometry (COG) is shifted by a distance $\textbf{r}_m$, a righting moment tend to restore the robot to its previous attitude after any rotational displacement. The moment can be modeled as 
\begin{align}
\label{equ:righting_moment}
    \mathbf{T}_m^{\text{head}} = mg \textbf{r}_m \sin{\beta},
\end{align}
where $\textbf{r}_m$ is displacement vector pointing from COM to COG, and $\beta$ is the pitch angle.

\textbf{Propulsive force from flagellum.} In Section \ref{sec:RSS}, we evaluate the hydrodynamics forces of each node along the discrete helical flagellum by the method of RSS. In the bi-flagellated system, two flagella provide the propulsion for the head. The propulsive force is equivalent to the resultant forces of all nodes forces of each flagellum applied on the two connection nodes $\mathbf{n}_{0}^1$ and $\mathbf{n}_{0}^2$, i.e., $\mathbf{f_p^\text{tail}} = \sum_{k=1}^{N} \mathbf{f_k}$.

For simplicity, we denote the resultant forces for two flagella as $\mathbf{f_p^1}$ and $\mathbf{f_p^2}$. Figure \ref{fig:RSS}(b) provides a time evolution of the resultant forces when two flagella rotate in the opposite direction. The force amplitude shows a sinusoidal pattern resulting from the long-ranged coupling between two flagella. The opposite sign of forces along the z-axis can cancel the propulsive effect but instead generate a turn-over torque because of the spacing distance $d$, which is the fundamental mechanism of the tumbling phenomenon. The torque applied equivalently on head node $\mathbf{n}^{\text{head}}$ is given by
\begin{align}
\label{equ:tail_moment}
    \mathbf{T}^\text{tail} =(\mathbf{f_p^1} - \mathbf{f_p^2}) \times \dfrac{d}{2},
\end{align}
where $\mathbf{r_{1}} = \mathbf{x}_{0}^1-\mathbf{x}^{\text{head}}$ and $\mathbf{r_{2}} = \mathbf{x}_{0}^2 - \mathbf{x}^{\text{head}}$.

In summary, the external forces and torques applied on the head includes $\mathbf{f}_{t}^{\text {head}}, \mathbf{T}_r^{\text{head}}, \mathbf{T}_m^{\text{head}}, \mathbf{f_p^\text{tail}}, \mathbf{T}^\text{tail}$. The governing equations of pivot steering in terms of pitch angle $\beta$ is:
\begin{equation}
\label{equ:pitch_dynamics}
    I_y \mathbf{\ddot{\beta}} = T_{rz}^{\text{head}} + T_{mz}^{\text{head}} + T^\text{tail}_z,
\end{equation}
where the subscript $z$ represents the torch component along z direction, and the governing equation of free swimming in terms of the DOF vector $\mathbf{Q}$ is
\begin{equation}
\begin{aligned}
    m\begin{bmatrix}
    0 \\
    \ddot{\mathbf{x}}^I
    \end{bmatrix}
    &= 
     \mathbf{q} \otimes 
     \begin{bmatrix}
     0 \\
     \mathbf{f}_{t}^{\text{head}}+\mathbf{f_p^1}+\mathbf{f_p^2}
     \end{bmatrix}
     \otimes \mathbf{q}^*, \\
    \dot{\mathbf{q}} &= \frac{1}{2} \mathbf{q} \otimes
    \begin{bmatrix}
    0 \\
    \mathbf{\omega}^B
    \end{bmatrix}, \\
    \mathbf{J} \dot{\mathbf{\omega}}^B &= -\mathbf{\omega}^B \times \mathbf{J} \mathbf{\omega}^B + \mathbf{T}_r^{\text{head}} + \mathbf{T}_m^{\text{head}} + \mathbf{T}^\text{tail},
\end{aligned}
\end{equation}
where $\mathbf{q}^*$ is the conjugate of $\mathbf{q}$, and $\mathbf{J} = \text{diag}(I_x, I_y, I_z)$ is matrix of moment of inertia.

\subsection{Control scheme of pivot tumbling}
\label{sec:control_scheme}
To study the controllability of tumbling, we rewrite Equation \ref{equ:pitch_dynamics} as the state space model by defining the state vector $\mathbf{x} \triangleq [\beta, \dot{\beta}]$ and control input vector $u \triangleq [\mathbf{f^1_p}, \mathbf{f^2_p}]$ (assuming small $\beta$, such that $\sin(\beta) \approx \beta$ holds):
\begin{equation*}
    \dot{\mathbf{x}}(t) = \mathbf{A}\mathbf{x}(t)+\mathbf{B}\mathbf{u}(t), \quad \mathbf{x(0)} = \mathbf{x}_0,
\end{equation*}
where 
\begin{equation*}
    \mathbf{A} = 
    \begin{bmatrix}
    0 & 1 \\
    -\dfrac{mgr_m}{I_y} & -\dfrac{8\pi C_r \mu h^3}{I_y}
    \end{bmatrix}
    , \mathbf{B} = \frac{d}{2I_y}
    \begin{bmatrix}
    0 & 0\\
    1 & -1
\end{bmatrix}
\end{equation*}

The system is structurally stable because the eigenvalues of matrix $\mathbf{A}$ have negative real part. The states of the system asymptotically converges to the steady condition
\begin{equation}
\label{equ:steady_condition}
\beta_\text{ss} = \dfrac{(\mathbf{f^1_p} - \mathbf{f^2_p}) d}{2 mg \textbf{r}_m}, \quad
\dot{\beta}_{\text{ss}} = 0
\end{equation}

Therefore, to realize the desired pitch angle $\beta_{\text{ref}}$, we require $\mathbf{f^1_p}$ and $\mathbf{f^2_p}$ to satisfy below conditions
\begin{equation}
\label{equ:condition}
\begin{aligned}
\mathbf{f^1_p}+ \mathbf{f^2_p} &= 0, \text{\quad (max torque)}\\
\dfrac{(\mathbf{f^1_p} - \mathbf{f^2_p}) d}{2 mg \textbf{r}_m} &= \beta_\text{ref}, \text{\quad (steady state value)} \\
\| \mathbf{f^1_p} \|, \| \mathbf{f^2_p} \| &\leq \mathbf{f_{\max}}. \text{\quad (effective propulsion)}
\end{aligned}
\end{equation}

However, to implement a actual control for the propulsion $\mathbf{f^1_p}$ and $\mathbf{f^2_p}$, we need more knowledge on the mechanism of flagellated propulsion. In Section \ref{sec:result}, we characterize the propulsion with flagellum geometry and rotation speed.

\subsection{Definition of the problem}
\label{sec:definition}

The general framework introduced above for the forced-head dynamics is now applied to generating bi-flagellated locomotion. We provide specifics on the geometry and physical parameters of this problem. The flagellum is chosen to be a rigid right-handed helical filament with Young's modulus $E$. The geometrical parameters describing the helix structure include helix pitch $\lambda$, helix radius $R$, axial length $l$, and cross-sectional radius $r_0$. The values of the parameters in Table \ref{tab:para} are chosen to match the laboratory experiments described in Section \ref{sec:experiment}. We use a dimensionless scheme to generalize the results, except for several fundamental variables, to make a valid comparison between macroscopic and microscopic mechanisms \cite{kim2003macroscopic}. The procedures are introduced as follows.

The helical flagellum is connected to the head at one extremity, where it is rotated counterclockwise with a prescribed angular velocity $\omega$. Two flagella are spaced with a specific distance $d$. Hereafter, we normalize the spacing distance by the helix radius, $R$, such that the normalized distance is $\Bar{d} = d/R$, where the overbar symbol $\Bar{\cdot}$ denotes the normalized variables. Likewise, the geometrical parameters that describes the flagellum structure include $\Bar{\lambda} = \lambda/l$ and $\Bar{R} = R/\lambda$. The input for the bi-flagellated system is angular velocity $\omega$ as a function of time. The propulsive thrust $F$ is the z component of $\mathbf{f_{p}^\text{tail}}$ and turnover torque $T$ is the y component of $\mathbf{T}^\text{tail}$, with the normalization as $\Bar{F} = F/(\mu \omega R L)$ and $\Bar{T} = T/(\mu \omega R^2 L)$, where $\mu$ denote the viscosity of fluid. This dimensionless representation allows for generality across length scales in interpreting our findings.

\section{Results and discussion}
\label{sec:result}
The bi-flagellated system can undergo different motions by varying the actuation modes of the two motors. In this section, we study the mechanism of tumbling behavior. We first explore the parameter space that enhances the direction change from flagellum geometry and robot structure. Then we show a bi-flagellated robot that can efficiently reorient the body by tumbling.


\subsection{Four distinct locomotion patterns}

\begin{figure}[ht!]
    \centering
    \includegraphics[width=\linewidth]{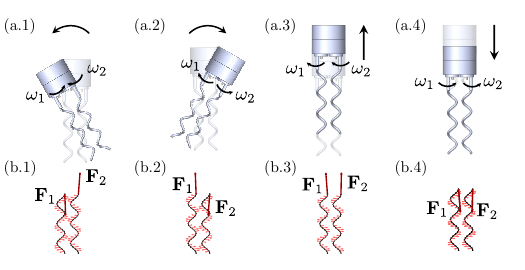}
    \caption{Locomotion patterns of the bi-flagellated robot. The robot can turn and translate when rotating two flagella with different modes. Denote $\omega > 0$ when flagellum rotating counterclockwise. (a.1) Left turn when $\omega_1>0, \omega_2 <0$, (a.2) Right turn when $\omega_1<0, \omega_2>0$, (a.3) upward translation when $\omega_1<0, \omega_2<0$, (a.4) downward translation when $\omega_1>0, \omega_2>0$. (b.1)-(b.4) The hydrodynamics force applied at two flagella with corresponding flagella rotation modes (a.1)-(a.4). The light red arrows represent the force direction applied on the node, and the dark red arrows represent the resultant force direction applied on the extremity of the flagellum.}
    \label{fig:motion_pattern}
\end{figure}

Since two flagella are rotated by two motors separately, we can propose four different actuation combinations according to the rotation direction of the two motors, i.e., 
\begin{enumerate}
    \item $\omega_1 > 0$, $\omega_2 < 0$ 
    \item $\omega_1 < 0$, $\omega_2 > 0$
    \item $\omega_1 < 0$, $\omega_2 < 0$
    \item $\omega_1 > 0$, $\omega_2 > 0$
\end{enumerate}
where we denote $\omega_i > 0$ as the counterclockwise rotation from the overlook view. The magnitude of angular velocities is equal regardless of the rotation direction, i.e., $|\omega_1| = |\omega_2|$. By the numerical simulation tool in Section \ref{sec:method}, we explore all the possible locomotion patterns from the above combinations. 

Figure \ref{fig:motion_pattern} (a) demonstrates four different locomotion induced by two flagella, including left turn, right turn, upward translation, and downward translation. The corresponding plots of the force field can interpret the dynamics mechanism behind each motion in Figure \ref{fig:motion_pattern} (b). For the right-handed helix, a counterclockwise rotation yield a resultant force pointing obliquely downward, while the clockwise rotation yield an upward force, as the dark red arrows show. Rotating two flagella with the same direction ensures the two resultant forces point up or down simultaneously but not precisely in the same direction due to the hydrodynamics coupling. On the contrary, rotating two flagella in the opposite direction make the resultant forces work as a force couple that is subject to generating a turnover torque. If the torque is large enough to overcome the intrinsic inertial of the head, directional change takes place with time evolution.

\begin{figure}[ht!]
    \centering
    \includegraphics[width=\linewidth]{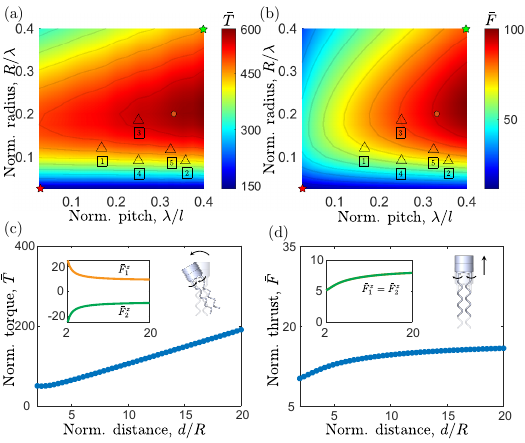}
    \caption{Geometrical parameters determine the magnitude of propulsive thrust and turnover torque. (a) Dependence of normalized propulsive force $\Bar{F}$ (log-scaled) on both normalized pitch $\lambda/l$ and normalized radius $R/ \lambda$. (b) Dependence of normalized turnover torque $\Bar{T}$ (log-scaled) on both normalized pitch $\lambda/l$ and normalized radius $R/ \lambda$. The upward triangle symbols in (a) and (b) correspond to Table \ref{tab:flagella}. The cycled red marker corresponds to the bi-flagellated robot. The red and green pentagram markers corresponds to $l = 1600 R$ and $l = 6.25 R$, respectively. (c) Normalized turnover torque $\Bar{T}$ as a function of normalized flagella spacing distance $d/R$. (d) Normalized propulsive thrust $\Bar{F}$ as a function of normalized flagella spacing distance $d/R$. Inset in (c) and (d): the normalized resultant force of two flagella as a function of normalized flagella spacing distance $d/R$. The non-linear relationship emerges when two flagella are in proximity, i.e., $d/R < 5$, due to hydrodynamics coupling between two flagella. }
    \label{fig:paraSpace}
\end{figure}

\begin{table}[ht]
  \centering
  \caption{Parameters of flagellum for five species of bacteria}
  \label{tab:flagella}
  \begin{tabular}{ cccc }
    \toprule
    Label & Microorganism & $\lambda/l$ & $R/\lambda$ \\  
    \midrule
    1 & Caulobacter crescentus \cite{koyasu1984caulobacter} & 0.1667 & 0.1205 \\ 
    \midrule
    2 & Escherichia coli \cite{darnton2007torque} & 0.3571 & 0.0909 \\  
    \midrule
    3 & Rhizobium lupini \cite{scharf2002real} & 0.2500 & 0.1852 \\    
    \midrule
    4 & Salmonella \cite{macnab1977normal} & 0.2500 & 0.0909 \\ 
    \midrule
    5 & Vibrio alginolyticus \cite{chattopadhyay2009effect} & 0.3243 & 0.1167 \\ 
    \midrule
    $\color{red} \circ$ & Bi-flagellated robot & 0.33 & 0.2 \\
    \bottomrule
  \end{tabular}
\end{table}

\subsection{Design space for optimal flagellum geometry}
\label{sec:geometry_dependence}

\begin{figure*}[ht!]
    \centering
    \vspace{-2ex}
    \includegraphics[width=\linewidth]{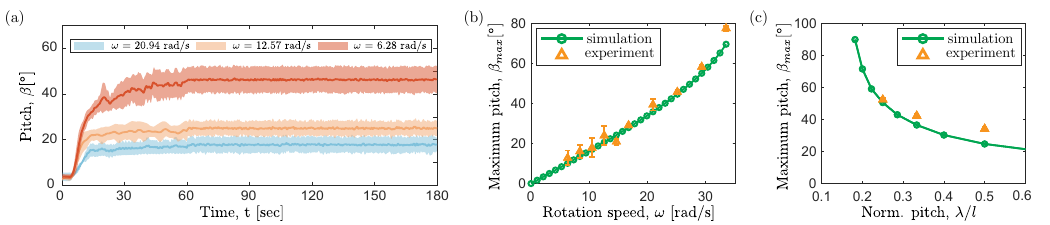}
    \vspace{-2ex}
    \caption{Comparison of simulation and experiment for fixed-body rotation. (a) Time evolution of pitch angle $\beta$ when angular speed of flagellum varying from 6.28 rad/s to 20.94 rad/s with $\lambda/l = 3, R/\lambda = 5, d/R = 3.5$. (b) The steady state values of pitch angle $\beta_{\text{max}}$, as a function of angular velocity $\omega$, with a parameter set: $\lambda/l = 3, R/\lambda = 5, d/R = 3.5$. (c) The steady state values of pitch angle $\beta_{\text{max}}$, as a function of normalized radius $R/\lambda$, with a parameter set: $l/R = 15, d/R = 3.5$, and $\omega$ = 20.9 rad/s. (d) The steady state values of pitch angle $\beta_{\text{max}}$, as a function of normalized pitch $R/\lambda$, with a parameter set: $R/\lambda = 5, d/R = 3.5$, and $\omega$ = 20.9 rad/s. }
    \label{fig:gimbalExp}
\end{figure*}

Thus far, our findings on the bi-flagellated actuation and corresponding locomotion patterns have brought insight into the run-and-turn behaviors of bacteria. Next, we perform a broader exploration of the parameter space for the structure of the helical flagellum and robot, emphasizing the ranges relevant to natural multi-flagellated cells. We use normalized propulsive thrust $\Bar{F}$ and turnover torque $\Bar{T}$ to represent propulsion and steering efficiency. $\Bar{F}$ characterize propulsion ability by a single helical flagellum. In contrast, $\Bar{T}$ represents the ability of direction change formed by the force couple accounting for the long-ranged hydrodynamics effect. A systematic parametric study is performed to quantify the dependence of these variables on the geometry and structure parameters.

In Figure \ref{fig:paraSpace} (a) and (b), we plot the magnitude of $\Bar{T}$ and $\Bar{F}$ (colorbar) in log-scale on the normalized pitch and normalized radius at $\omega = 20.94$ rad/s. We set helix radius $R$ as a constant value of 0.0064 m and compute the helix pitch $\lambda$ and length $l$ per the corresponding value of normalized values. In the plot of torque, we make the distance between two flagella as constant $d = 3 R$ to eliminate the effect of spacing distance $d$. Two phase diagrams have commonalities in the geometrical parameters. We find that the two quantities increases as the normalized pitch and radius increase. The highest value locate at the left-bottom region, indicating the maximum torque and thrust. However, the region represents an elongated flagellum concerning the radius, i.e., $L = 1600 R$, which is less common in bacteria due to its poor maneuverability and high energy cost. From the parameter distribution of several species of bacteria in Table \ref{tab:flagella}, as the upward triangles, we learn that the optimal geometry locates on the right-bottom region is a trade-off between dynamic performance and dimensionality. Therefore, as the red marker shows, we take $\lambda/l = 0.33$ and $R/\lambda = 0.2$ as representative values.

In Figure \ref{fig:paraSpace} (c) and (d), we plot the magnitude of $\Bar{T}$ and $\Bar{F}$ as a function of normalized space distance $d/R$, at the representative geometrical values. As $d/R$ increases, $\Bar{T}$ and $\Bar{F}$ monotonically increase, but non-linearity of the curve due to hydrodynamics occurs when two flagellum is in proximity. Intriguingly, from the insets of two plots, we see that the long-ranged hydrodynamics exert different effects when two flagella rotate differently. Hydrodynamics escalates the force magnitude when they rotate in the opposite direction but decreases the magnitude when they rotate in the same direction. Given torque is a product of force and arm, the magnitude almost remains the same when the normalized spacing distance $d/R$ increases from 2 to 3.5. The result implies the maximum propulsive force and turnover torque occur when two flagella are spaced an infinite distance. However, the long spacing distance is not preferable for either the bacteria or bio-inspired robot. The multi-flagellated system must compromise with dimension and propulsion efficiency. Here, we take $d/R = 3.5$ in our bi-flagellated system to ensure proper size of the head.

\subsection{Analysis on tumbling process}
Toward validating the numerical simulations presented in Section \ref{sec:method}, we now perform a direct quantitative comparison with experimental results using the apparatus described in Section \ref{sec:experiment}. Emphases are given to the evolution of pitch angle $\beta$, as an accumulative result of applied forces and torques. 

In this section, we investigate how pitch angle $\beta$ evolved with different flagellum parameters, including normalized pitch $\lambda/l$ and rotation speed $\omega$. In Figure \ref{fig:gimbalExp}(a), we plot $\beta$ as a function of time $t$, with the reference assumed $\lambda/l = 3, R/\lambda = 5, d/R = 3.5$. The pitch angle keeps increasing with time and reaches a maximum value, denoted as $\beta_\text{max}$. The magnitude of $\beta_{max}$ is proportional to the rotation speed of the flagella, which is ensured by the steady condition in Equation \ref{equ:steady_condition}. Therefore, though without measurement the value of turnover torque in experiment, we can use the $\beta_\text{max}$, a.k.a., $\beta_\text{ss}$ as an indicator of torque magnitude. Figure \ref{fig:gimbalExp} (b) plot the relationship between $\beta_{\text{ss}}$ and rotation speed $\omega$, and show an excellent agreement between experiments and simulations. We learn that the magnitude of torque generated by forces $\mathbf{f^{1}_p}$ and $\mathbf{f^{2}_p}$ is linear with the rotation speed of the flagellum.

We employ our numerical simulations to explore the effect of $\lambda/l$ with a comparison with experiments when keeping the angular velocity of flagellum $\omega = 20.94$ rad/s, for which we showed in the previous results that there is a significant direction change effect. Figure \ref{fig:gimbalExp}(c) show a good match between the experiment and simulation of the tendency of $\beta_\text{max}$ on $\lambda/l$. The agreement validate the result in Figure \ref{fig:paraSpace}(a) about the relationship between turnover torque and flagellum geometry.

\subsection{Attitude control of bi-flagellated robot}
\begin{figure}[ht!]
    \centering
    \includegraphics[width=\linewidth]{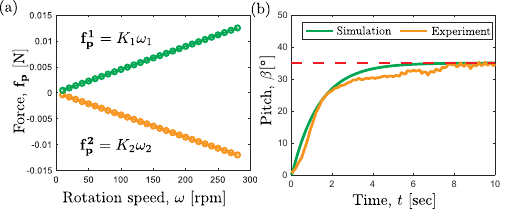}
    \caption{Attitude control of bi-flagellated robot. (a) Relationship between propulsion force and flagellum rotation speed $\omega$ with $d/R = 0.35, R/\lambda = 0.2$, and $\lambda/l = 0.33$. The slopes $K_1 = 0.0006$ Ns/rad and $K_2 = -0.0006$ Ns/rad. (b) The evolution of pitch angle $\beta$ when apply the attitude control scheme with a reference $\beta_{\text{ref}} = 35^{\circ}$.}
    \label{fig:control}
\end{figure}
The previous sections shows that the propulsion force and turnover torque are associated with flagellum structure $d/R, R/\lambda, \lambda/l$ and rotation speed $\omega$. As for the control problem, it is not feasible to change the structure related parameters to vary the propulsion force and torque in process. Therefore, we take the rotation speed of two flagella $\omega_1, \omega_2$ as the actual control variable. Figure \ref{fig:gimbalExp}(b) illustrate the torque is linear with the rotation speed for given flagellum structure. Through our computation framework, we evaluate the propulsion force by $\mathbf{f^1_p} = K_1 \omega_1, \mathbf{f^2_p} = K_2 \omega_2$ as Figure \ref{fig:control}(a). This allows us to realize the control scheme mentioned in Section \ref{sec:control_scheme}. We showcase a tracking for a constant attitude angle $\beta_{\text{ref}} = 35^{\circ}$ in both simulation and experiment. By solving Equation \ref{equ:condition}, we obtain $\omega_1 = 201.6 \text{rpm}, \omega_2 = -201.6 \text{rpm}$. Then we set the rotation speed of two flagellum with the values and the pitch angle $\beta$ evolved as Figure \ref{fig:control}(b).

\section{Conclusions and future work}
\label{sec:conclusion}
In conclusion, we present a bi-flagellated mechanism and a numerical simulation framework for studying bacteria tumbling behavior. A dimensionless scheme generalize our results to the bacteria level. The framework are used to explore the relationship between the steering ability and the structural parameters of the bi-flagellated system. The attitude control scheme ensure us to control the orientation of the robot.

Directions for future work include: (i) formulation of optimal control policy for free swimming robot, and (ii) develop the simulation tools for soft and elastic flagellum, accounting for the contact effect between flagella.

\end{document}